\documentclass[conference]{IEEEtran}
\IEEEoverridecommandlockouts

\usepackage{cite}
\usepackage{amsmath,amssymb,amsfonts}
\usepackage{algorithmic}
\usepackage{graphicx}
\usepackage{textcomp}
\usepackage{xcolor}
\usepackage{hyperref}
\usepackage{verbatim}
\usepackage{multirow}
\usepackage{rotating}
\usepackage{tabularray}
\usepackage{makecell}
\usepackage{booktabs}
\usepackage{tikz}
\usepackage{circuitikz}
\usepackage{blindtext}
\UseTblrLibrary{siunitx}

\def\BibTeX{{\rm B\kern-.05em{\sc i\kern-.025em b}\kern-.08em
    T\kern-.1667em\lower.7ex\hbox{E}\kern-.125emX}}

\begin{document}

\title{GPT Struct Me: Probing GPT Models \\ on Narrative Entity Extraction}

 \author{\IEEEauthorblockN{Hugo Sousa\textsuperscript{\textsection}}
 \IEEEauthorblockA{\textit{INESC TEC} \\
 \textit{University of Porto}\\
 Porto, Portugal \\
 hugo.o.sousa@inesctec.pt}
 \and
 \IEEEauthorblockN{Nuno Guimarães\textsuperscript{\textsection}}
 \IEEEauthorblockA{\textit{INESC TEC} \\
 \textit{University of Porto}\\
 Porto, Portugal \\
 nuno.r.guimaraes@inesctec.pt}
 \and
 \IEEEauthorblockN{Alípio Jorge}
 \IEEEauthorblockA{\textit{INESC TEC} \\
 \textit{University of Porto}\\
 Porto, Portugal \\
 alipio.jorge@inesctec.pt}
 \and
 \IEEEauthorblockN{Ricardo Campos}
 \IEEEauthorblockA{\textit{INESC TEC} \\
 \textit{University of Beira Interior}\\
 Porto, Portugal \\
 ricardo.campos@inesctec.pt}
 }

\maketitle

\begingroup\renewcommand\thefootnote{\textsection}
\footnotetext{Equal contribution.}
\endgroup

\begin{abstract}
    The importance of systems that can extract structured information from textual data becomes increasingly pronounced given the ever-increasing volume of text produced on a daily basis. Having a system that can effectively extract such information in an interoperable manner would be an asset for several domains, be it finance, health, or legal. Recent developments in natural language processing led to the production of powerful language models that can, to some degree, mimic human intelligence. Such effectiveness raises a pertinent question: Can these models be leveraged for the extraction of structured information? In this work, we address this question by evaluating the capabilities of two state-of-the-art language models -- GPT-3 and GPT-3.5, commonly known as ChatGPT -- in the extraction of narrative entities, namely events, participants, and temporal expressions. This study is conducted on the Text2Story Lusa dataset, a collection of 119 Portuguese news articles whose annotation framework includes a set of entity structures along with several tags and attribute values. We first select the best prompt template through an ablation study over prompt components that provide varying degrees of information on a subset of documents of the dataset. Subsequently, we use the best templates to evaluate the effectiveness of the models on the remaining documents. The results obtained indicate that GPT models are competitive with out-of-the-box baseline systems, presenting an all-in-one alternative for practitioners with limited resources. By studying the strengths and limitations of these models in the context of information extraction, we offer insights that can guide future improvements and avenues to explore in this field.    
\end{abstract}


\section{Introduction}
A narrative is a type of written or spoken discourse that tells a story by describing a sequence of connected events~\cite{Toolan2012Narrative:Introduction}. At its core, a narrative is composed of three main entities: participants, events, and temporal expressions (referred as timexs henceforward). Participants are the drivers of the actions, and can be people, animals, or abstract entities. Events refer to what happens to these participants, forming the plot of the narrative. Lastly, timexs help establish the sequence, duration, and frequency of these events, providing the timeline of the narrative. Given the prevalence of narratives in human communication, methods for the automatic extraction of narrative entities have garnered significant attention and research efforts from the natural language processing (NLP) community~\cite{Santana2023AData}. As an example, in the narrative \textit{``The suspect was convicted last night.''}, a narrative entity extraction system must be able to identify \textit{``the suspect''} as a participant, \textit{``convicted''} as an event, and \textit{``last night''} as a timex. 

The field of NLP has evolved significantly in recent years, mainly driven by the introduction of the Transformer architecture~\cite{Vaswani2017AttentionNeed}. This architecture has been used in a multitude of large language models (LLMs) such as BERT~\cite{Devlin2019BERT:Understanding}, and BLOOM~\cite{Workshop2022BLOOM:Model}. Initially, the main paradigm was to apply these models to downstream tasks by fine-tuning them. However, the introduction of GPT-3~\cite{Brown2020LanguageLearners} led to a shift towards ``prompt-based learning'', emphasizing the optimization of the input prompt rather than the model itself~\cite{Liu2023Pre-trainProcessing}. This paradigm was further reinforced with the introduction of GPT-3.5, commonly known as ChatGPT~\cite{OpenAIIntroducingChatGPT}. Such advances have led several research studies to probe the capabilities of LLMs such as reasoning~\cite{Yao2022ReAct:Models}, semantics~\cite{ZhangProbingTasks}, machine translation\cite{Hendy2023HowEvaluation}, and named entity recognition~\cite{Qin2023IsSolver}, presenting state-of-the-art results in some of them. 

The rising capabilities of these models present an intriguing question: Can LLMs effectively extract narrative entities from textual data? In an effort to answer this question we probe, GPT-3 and ChatGPT on the extraction of events, participants, and timexs on a set of news articles written in European Portuguese. To design the prompts, we conducted experiments using various templates on a subset of $20$ carefully sampled and representative documents of our dataset. After identifying the best prompt templates for each entity-model pair, we assess GPT-3 and ChatGPT effectiveness and compared them with out-of-the-box baseline systems in a partition of the data that we allocated to test the models.

In summary, the main contributions of this paper are:

\begin{enumerate}
    \item A prompt engineering methodology to evaluate LLMs using different levels of information to leverage their capabilities in the extraction of entities in narrative texts.

    \item A robust evaluation of GPT-3 and ChatGPT in the extraction of narrative entities in European Portuguese news.
\end{enumerate}

The remainder of the paper is structured as follows: Section~\ref{sec:soa} frames our research within the context of existing studies. Section~\ref{sec:methods} provides a description of the experimental setup employed for this research. Subsequently, Section~\ref{sec:prompt} briefly outlines the construction of the prompts along with the preliminary ablation study performed for the selection of the most effective prompt-model-entity triplet. Section~\ref{sec:evaluation} presents the results of the GPT models and compares these with the results achieved by baseline systems. Lastly, Section~\ref{sec:conclusion} presents the conclusions and sheds light on potential avenues for future research. 
\section{Related Work}\label{sec:soa}
The Generative Pre-trained Transformers (GPT) models from OpenAI\cite{Brown2020LanguageLearners,OpenAIIntroducingChatGPT} presented several leaps in effectiveness, achieving state-of-the-art results in a variety of NLP tasks~\cite{Srivastava2022BeyondModels}, and demonstrated impressive capabilities across a broad spectrum~\cite{Qin2023IsSolver}. For instance, in machine translation, GPT models achieved competitive effectiveness when compared with domain-specific state-of-the-art systems in high-resource languages~\cite{Hendy2023HowEvaluation}. Additionally, GPT-3 summaries have been preferred by human evaluators in text summarization tasks~\cite{Goyal2022NewsGPT-3}. 

Regarding the capabilities of LLMs in extracting structured information from text, the research studies present mixed results. Tang et al.~\cite{Tang2023DoesMining} reported poor effectiveness from ChatGPT in the named entity recognition task in the medical domain, achieving an $F_1$ score of only $23.37\%$. Conversely, in the narrative domain~\cite{Toolan2012Narrative:Introduction}, Stammbach et al.~\cite{Stammbach2022HeroesData} found that GPT-3 outperformed previous methods in extracting participants, from narrative texts using a zero-shot prompt approach. 

For the extraction of events, there have been works~\cite{Du2020EventQuestions,Liu2020EventComprehension} that achieved state-of-the-art results on the ACE dataset~\cite{LinguisticDataConsortiumACEEntities} by treating event extraction as a machine reading comprehension task. More recently, GPT models have also been evaluated for the same task~\cite{Gao2023ExploringExtraction}. Nevertheless, the results indicated that ChatGPT is not robust enough. Furthermore, the authors also stated that the refinement of the prompt did not improve the effectiveness of the models. However, the study used a different prompt template from ours. Finally, concerning temporal expression extraction, our research fills a gap since, to the best of our knowledge, no studies have explored prompt-based approaches in this domain.
\section{Experimental Setup}\label{sec:methods}

With this study, we intend to probe the capabilities of LLMs in the extraction of narrative entities through a prompt-based approach. We aim to explore how well these models can generate and comprehend narrative content by leveraging their vast language understanding and generation abilities. 

To ensure the replicability of the experiments, we made the code available (as well as more details on the experimental setup and results) in a GitHub repository\footnote{\url{https://github.com/hmosousa/gpt_struct_me}}.

\subsection{Model Selection}

When selecting which models to implement in our research, several layers of criteria had to be taken into account.

\paragraph{\textbf{Effectiveness}} We aim to apply models that present state-of-the-art results on the generative task. This ensures that our research is not only up-to-date but also contributes with new findings to the scientific community.

\paragraph{\textbf{Hardware Requirements}} The most effective LLMs are (typically) the ones with the highest number of parameters. This implies a demand for specific hardware that can execute these large models, which is far beyond the capacity of typical consumer-grade equipment.

\paragraph{\textbf{Availability}} Although numerous LLMs have been developed, most of them are typically owned and maintained by large corporations or research institutions that choose to keep them proprietary. This includes models such as PaLM~\cite{Chowdhery2022PaLM:Pathways} and Chinchilla~\cite{Hoffmann2022TrainingModels}, which despite reporting great effectiveness on several benchmarks, are not of public access, limiting their study.

\paragraph{\textbf{Domain}} We sought models that have been developed for general use, rather than being fine-tuned for a specific task. For example, CodeX~\cite{Chen2021EvaluatingCode} has been optimized for producing code, and Galatica~\cite{Taylor2022Galactica:Science} has been optimized for scientific texts. Such specialization was not desirable for our research. 

After evaluating all these factors, GPT-3 and ChatGPT emerged as the obvious choices. These models are not only effective and easily accessible but also suitable for a broad range of tasks, which aligns with the objectives of our research.  

However, it is important to acknowledge certain drawbacks associated with these models. The most prominent is the lack of reproducibility and control, as the OpenAI API\footnote{\url{https://openai.com/blog/openai-api}} parameters alone cannot guarantee deterministic answers. To address this concern, we set the \texttt{\small temperature} parameter of the API to zero. Although this does not ensure complete determinism, it significantly reduces variability\footnote{\url{https://platform.openai.com/docs/guides/gpt/why-are-model-outputs-inconsistent}}. Despite this limitation, using these models provides us with a reliable estimate of the effectiveness of current state-of-the-art LLMs, which is the main goal of this study.

\subsection{Dataset Description}

Three main factors guided the selection of the appropriate dataset. The first concern was to ensure that the dataset had not been employed during the training phase of the models. This factor mitigates the risk of data leakage which could artificially inflate the results. The second requirement was that the dataset included annotations for all the entities, namely: events, participants, and timexs. The final consideration was the dataset's size and complexity since it had to be sufficiently large and diverse to yield trustable results.

A dataset we found to meet all the above criteria is the Text2Story Lusa dataset~\cite{Silvano2023Text2StoryAnnotated}. This dataset comprises $119$ news articles written in European Portuguese that were densely annotated with a multilayer semantic annotation scheme~\cite{Silvano2021DevelopingCorpus}, which include the annotation of $3,057$ events, $3,546$ participants, and $325$ timexs. Furthermore, this dataset was only released after the development of this research project, therefore ensuring that the models did not have access to it in the learning phase.

\subsection{Task Definition}

Conventional methods treat narrative extraction as an extractive task. This involves having the system pinpoint the exact position or ``offset'' of tokens that refer to an entity within a given passage. However, when humans are asked to identify entities within a text, they do not pinpoint the exact offsets of those entities within the text. Instead, humans tend to list out the entities they identify. For instance, if one was asked to name the characters in a story, the answer would likely be the list of entities -- ``Mary'', ``Gaspar'', ``the baby'' -- rather than providing the offsets of their mentions within the text.

In our work, we follow this human-oriented approach, by prompting the models to list the entities in the text rather than presenting the offsets in which the entities appeared. Then we compared this list of predicted entities with the list of entities on the annotation file as explained in the next section.

\subsection{Evaluation Metrics}
To evaluate the effectiveness of the systems we compare the annotated and model-generated entity lists with the traditional precision ($P$), recall ($R$), and $F_1$ score. These, however, only take into account exact matches as correct, ignoring predictions that partially overlap with the true label. Thus, to enhance the comprehensiveness of our evaluation, we introduce a supplementary metric: the relaxed $F_1$ score ($F_{1_{r}}$). This score considers the partial (but contiguous) overlap between the annotated entities and the ones predicted by the system, thereby accounting for a more nuanced evaluation.

\section{Prompt Engineering}\label{sec:prompt}

Extracting the best effectiveness of generative models relies heavily on carefully constructing the prompt. Although several studies have been developed~\cite{Liu2023Pre-trainProcessing}, the process of constructing an effective prompt is iterative and experimental, requiring multiple iterations and empirical experiments to achieve the desired outcomes. 

\subsection{Prompt Template Definition}

We crafted a prompt template that we found to be able to successfully complete the narrative entity extraction task. This is composed of seven modules \footnote{We encourage interested readers to delve into our code repository, where they can find not only the code responsible for generating the prompts but also the generated prompts.}:

\begin{enumerate}
    \item{\textbf{Task}} States the task that the generative model is expected to perform.
    
    \item{\textbf{Definition}} The description associated with the entity being extracted. The idea is to provide the model with the same guidelines given to the annotators.
    
    \item{\textbf{Classes}} A list of possible classes for the entity being extracted. However, when using this module, we also modify the task of the model to ``Extract and classify all \texttt{\small <entity>}.'' instead of just ``Extract all \texttt{\small <entity>}.''. The idea is that by asking the models to perform a classification task we are encouraging them to gain a more nuanced understanding of the text content, as they need to not only identify the specific entity but also understand and categorize it appropriately. This aligns, to some extent, with the chain-of-though prompting approach, which has been shown to improve the ability of LLMs to perform complex reasoning~\cite{Wei2022Chain-of-ThoughtModels}.
    
    \item{\textbf{Example}} This is a simple one-shot example that presents the input text along with the expected output. 

    \item{\textbf{Format}} To ensure that the generated outputs could be programmatically parsed, we specified the required format explicitly. 
    
    \item{\textbf{Input}} The input text from which we want the model to extract the entities.
    
    \item{\textbf{Output}} This is the last line of the prompt, left blank for the model to generate the answer.
    
\end{enumerate}

Note that some of the components in the template are not required for the task, namely: Definition, Classes, and Example. The inclusion of these aims to provide additional information to the prompt, potentially enhancing the effectiveness of the model. However, since we cannot determine \textit{a priori} which of these modules is beneficial, we conduct a preliminary ablation study to evaluate the impact of each optional module of the template on each model-entity pair.

\subsection{Preliminary Ablation Study}

We use $20$ documents from our dataset to estimate the effectiveness of each prompt-model-entity triplet. Due to the small size of the sample, one needs to be careful in the sampling process, since the goal is to have a sample that produces a good estimate of the expected effectiveness of the templates on the full corpus. Thus, the following steps were used to select the documents to be included in the sample: 1) Compute the $T-1$ quantiles of the number of tokens in the documents, with $T$ being the number of documents to be sampled (in our case $T=20$); 2) Divide the dataset into $T$ ranges, each being bounded by the computed quantiles; and 3) Randomly sample one document from each range. Using this procedure, we ensured that the sample was representative of the entire corpus. Additional information regarding the sampling process is available in our code repository. 

After building the prompts for each of the $20$ documents, we prompt the models. We briefly summarize the results achieved.

First, the effectiveness of both models significantly differs according to the type of entity. The GPT models' capability in extracting timexs seems to achieve better results when compared to participants and events. When comparing GPT-3 and ChatGPT, the first achieves a better overall effectiveness in the extraction of timexs and participants while ChatGPT seems to be better in the extraction of events. Second, when analyzing the results at a prompt level, we observed that adding more information does not always translate into better results (for example, combining the Definition and the Task results in an overall worse effectiveness than prompting only with the Task). Third, the inclusion of the Example improves, on average, the $F_1$ score in the different extraction tasks. Lastly, one-shot prompts provide the best results when they combine all three types of information, while in zero-shot prompts the best results are achieved when the prompts do not include Definitions. A more detailed description of the results is available in the paper repository.

Given the results, we select the six prompts that achieved the highest $F_1$ score for each entity-model pair and use them to evaluate the models' effectiveness on the remaining documents of the dataset.

\section{Model Evaluation}\label{sec:evaluation}
For the final evaluation, the remaining $96$ documents within the dataset were used. The GPT models' effectiveness is compared against two established baselines for the extraction of events (SRL ~\cite{Oliveira2021ImprovingLearning} and TEFE ~\cite{Sacramento2021JointLanguage}) and timexs (HeidelTime ~\cite{Strotgen2013MultilingualTagging} and TEI2GO ~\cite{Sousa2023TEI2GO:Identification}), and one baseline for the participant extraction task (SRL ~\cite{Oliveira2021ImprovingLearning}). The results are presented in Table~\ref{tab:results}. 

\begin{table}
    \caption{Results on the test set of the Baseline and GPT models.}\label{tab:results}
    \centering
    \refstepcounter{table}
    \begin{tblr}{
      colspec = {X[2]X[2]X[si={table-format=2.1},c]X[si={table-format=2.1},c]X[si={table-format=2.1},c]X[si={table-format=2.1},c]},
      cell{2}{1} = {r=5}{},
      cell{7}{1} = {r=3}{},
      cell{10}{1} = {r=4}{},
      hline{1,14} = {-}{0.08em},
      hline{2,7,10} = {-}{0.05em},
      hline{5,8,12} = {-}{dotted},
    }
                          &             & {{{$P$}}} & {{{$R$}}} & {{{$F_{1}$}}} & {{{$F_{1_{r}}$}}} \\
    \textbf{Timexs}       & TEI2GO      & 50.7      & 59.4      & 54.7          & 55.7              \\
                          & HeidelTime  & 50.1      & 59.3      & 54.2          & 56.3              \\
                          & SRL         & 23.6      & 35.7      & 28.4          & 50.5              \\
                          & GPT-3       & 58.5      & 73.5      & \textbf{65.1} & \textbf{73.0}     \\
                          & ChatGPT     & 34.5      & 58.6      & 43.5          & 48.0              \\
    \textbf{Participants} & SRL         & 27.1      & 20.7      & 23.4          & 47.7              \\
                          & GPT-3       & 35.9      & 35.8      & 35.8          & \textbf{51.0}     \\
                          & ChatGPT     & 40.1      & 46.6      & \textbf{43.1} & 47.4              \\
    \textbf{Events}       & SRL         & 67.1      & 40.4      & \textbf{50.4} & 83.5              \\
                          & TEFE        & 95.4      & 9.1       & 16.6          & \textbf{95.6}     \\
                          & GPT-3       & 41.9      & 53.3      & 46.9          & 51.7              \\
                          & ChatGPT     & 50.7      & 44.3      & 47.2          & 61.4              \\
    \end{tblr}
\end{table}

As observed in Table ~\ref{tab:results}, the GPT models outperform the baselines in the extraction of timexs and participants. However, concerning events, we can see that SRL and TEFE beat both GPT models in the $F_1$ strict and relaxed metrics. Given that all the baselines are used out-of-the-box while the prompts used in one-shot include an annotation example, the results go according to what was expected in timexs and participants. However, it is interesting to see that both GPT models cannot surpass the baselines in the events extraction task. We hypothesize that, given that the concept of event in the annotation is very specific, one annotation example is not enough for these models. Since SRL and TEFE are more task-driven models and are already more familiar with the concept through their learning process, they can outperform more broader and general models without seeing an example of the data beforehand.

In the extraction of timexs and participants, the GPT models outperform the best baseline in $10$ and $20$ points of strict $F_1$, respectively. We hypothesize that since the concept of timexs and participants in our annotation is more aligned with the common knowledge, GPT models have a better chance to outperform the baselines due to the large textual data on which they were trained on. In addition, with the inclusion of an example, the model is capable of gathering additional knowledge making it capable of surpassing the results achieved by the baselines. 

Finally, we can see that ChatGPT outperforms GPT-3 in the extraction of participants and events, with GPT-3 having better effectiveness in timexs. However, it is important to note that our evaluation is not focused on the models themselves but on the combination of these with our prompt methodology. We provide a uniform approach that enables the results presented in Table~\ref{tab:results} but, by itself, it is not guaranteed to maximize the effectiveness of GPT models.  

\section{Conclusions and Future Work}\label{sec:conclusion}

In this work, we evaluate through prompt-based learning the capability of GPT-3 and ChatGPT in the task of narrative entity extraction. To ensure a systematic evaluation, we develop a prompt methodology that combines different levels of knowledge. We conduct a preliminary ablation study to choose the best template prompts for each model-entity pair on a subset of the dataset and use the best templates to compare the models with a set of baselines on the remaining documents of the dataset. The results show that the combination of the best prompts and GPT models outperform (using strict $F_1$) the baseline in the extraction of participants and timexs but failed to reach the same results in the extraction of events. Although there is no definitive guarantee that superior results could not be attained using alternative prompt templates, the robustness of out-of-the-box models, applicable to a variety of entity extraction tasks, still offer significant advantages since they do not require exhaustive experimentation with different prompts. 

In future work, we intend to refine and improve our prompt methodology and evaluate more complex tasks in narrative extraction using LLMs, like the extraction of temporal, aspectual, objectual, and semantic role links. In addition, we also intend to tweak the current prompts, for instance, by extending the number of examples used on the Example component of the prompt. Further, we intend to access the present approach in other 
languages and domains to extend the evaluation of the models. All these possibilities combined with the release of more effective LLMs (such as GPT-4~\cite{OpenAI2023GPT-4Report}) lead to several promising research paths on narrative extraction that are worth exploring.

\section{Acknowledgments}

This work is financed by National Funds through the Portuguese funding agency, FCT - Fundação para a Ciência e a Tecnologia, within project LA/P/0063/2020 and the project StorySense, with reference 2022.09312.PTDC.

\bibliographystyle{IEEEtran}
\bibliography{references}

\end{document}